\documentclass[10pt]{iopart}

\usepackage{iopams}  
\usepackage{graphics} 
\expandafter\let\csname equation*\endcsname\relax
\expandafter\let\csname endequation*\endcsname\relax
\usepackage{amsmath} 
\usepackage{amssymb}  
\usepackage{siunitx}
\usepackage{nicefrac}
\usepackage{multicol}
\usepackage{multirow}
\usepackage{units}
\usepackage{array}
\usepackage{pgfplots} 
\usepackage{kantlipsum}
\usepackage{setstack}
\pgfplotsset{compat=1.14}
\usepackage[shortlabels]{enumitem}

\begin{document}

\title[Proprioceptive sensing and feedback]{\LARGE \bf  Effective Locomotion at Multiple Stride Frequencies Using Proprioceptive Feedback on a Legged Microrobot}
   \author{Neel~Doshi$^{1,2,}\ddagger$, Kaushik~Jayaram$^{1,2,}\footnote{These authors contributed equally}$, Samantha Castellanos$^{1,2}$, Scott Kuindersma$^{1}$ and Robert~J.~Wood$^{1,2}$}
\address{$^1$ John A. Paulson School of Engineering and Applied Sciences, Harvard University, Cambridge, MA, USA}
\address{$^2$Wyss Institute for Biologically Inspired Engineering, Cambridge, MA, USA}
\eads{\mailto{ndoshi}, \mailto{kjayaram @seas.harvard.edu}}

\begin{abstract}
Limitations in actuation, sensing, and computation have forced small legged robots to rely on carefully tuned, mechanically mediated leg trajectories for effective locomotion. Recent advances in manufacturing, however, have enabled the development of small legged robots capable of operation at multiple stride frequencies using multi-degree-of-freedom leg trajectories. Proprioceptive sensing and control is key to extending the capabilities of these robots to a broad range of operating conditions. In this work, we use concomitant sensing for piezoelectric actuation with a computationally efficient framework for estimation and control of leg trajectories on a quadrupedal microrobot. We demonstrate accurate position estimation ($<$\SI{16}{\%} root-mean-square error) and control ($<$\SI{16}{\%} root-mean-square tracking error) during locomotion across a wide range of stride frequencies (\SIrange{10}{50}{\hertz}). This capability enables the exploration of two blue bioinspired parametric leg trajectories designed to reduce leg slip and increase locomotion performance (e.g., speed, cost-of-transport, etc.). Using this approach, we demonstrate high performance locomotion at stride frequencies of (\SIrange{10}{30}{\hertz}) where the robot's natural dynamics result in poor open-loop locomotion. Furthermore, we validate the biological hypotheses that inspired the our trajectories and identify regions of highly dynamic locomotion, low cost-of-transport (3.33), and minimal leg slippage ($<$ 10 \%).
\end{abstract}


\vspace{2pc}
\noindent{\it Keywords}: self-sensing, linear-quadratic-gaussian control, legged microrobots, piezoelectric actuation, robust locomotion

\submitto{\BB}
%
\maketitle
%
\ioptwocol


\section{Introduction}

Terrestrial animals use a variety of complex leg trajectories to navigate natural terrains \cite{manton1977arthropoda}. The choice of leg trajectory is often determined by a combination of morphological factors including posture \cite{gatesy1991bipedal}, hip and leg kinematics \cite{geyer2006compliant}, ankle and foot designs \cite{kenning2017ultimate}, and actuation capabilities (e.g., muscle mechanics \cite{daley2003muscle, birn2014don}). In addition, animals also modify their leg trajectories to meet performance requirements such as speed \cite{mcmahon1985role}, stability \cite{daley2006running, wilshin2017longitudinal}, and economy \cite{dickinson2000animals}, as well as to adapt to external factors such as terrain type \cite{gorb1995design, jayaram2018transition} and surface properties \cite{yang2016nature, jayaram2016cockroaches}. 

Inspired by their biological counterparts, large (body length (BL) $\sim$\SI{100}{\centi\meter}) bipedal \cite{sakagami2002intelligent, hubicki2016atrias} and quadrupedal  \cite{raibert2008bigdog,seok2015design, tmech16semini, hutter2012starleth} robots typically have two or more actuated degrees-of-freedom (DOF) per leg to enable complex leg trajectories. This dexterity is leveraged in a variety of control schemes to adapt to different environments and performance requirements. For example, optimization algorithms have been used to command leg trajectories to enable stable, dynamic locomotion on the Atlas bipedal \cite{kuindersma2016optimization} and HyQ quadrupedal \cite{tmech16semini} robots. Furthermore, the MIT Cheetah \cite{seok2015design} relies on a hierarchical control scheme where the low-level controllers alter leg trajectories to directly modulate ground reaction forces.

However, as the robot's size decreases, manufacturing and material limitations constrain the number of actuators and sensors. Consequently, a majority of medium (BL $\sim$\SI{10}{\centi\meter}) \cite{saranli2001rhex} and small (BL $\sim$\SI{1}{\centi\meter}) \cite{birkmeyer2009dash, pierre2017gait, dharmawan2018design} legged robots have at most single DOF legs driven by a hip actuator. In such systems, leg trajectory is dictated by the transmission design, and these robots often rely on tuned passive dynamics to achieve efficient locomotion  \cite{bailey2001comparing, kim2006isprawl}. Nevertheless, careful mechanical design allows these robots to demonstrate impressive capabilities, including high-speed running  \cite{haldane2015running}, jumping \cite{haldane2018robotic}, climbing \cite{birkmeyer2011clash,birkmeyer2012dynamic}, horizontal to vertical transitions \cite{jayaram2018transition}, and confined space locomotion \cite{jayaram2016cockroaches}.

Recent work has also focused on developing whole-body locomotion control schemes for the autonomous operation of these small legged robots. These include controllers designed using stochastic kinematic models on the octopedal OctoRoACH \cite{karydis2015probabilistically} and using deep reinforcement learning on the hexapedal VelociRoACH \cite{nagabandi2018learning} robots. However, these robots do not have the mechanical dexterity to actively vary the shape of the their leg trajectory and instead rely on mechanical tuning and inter-leg timing (i.e., gait) to achieve effective locomotion at a specific operating frequency. 

\begin{figure}[t]
	\begin{center}
		\includegraphics[width=\columnwidth]{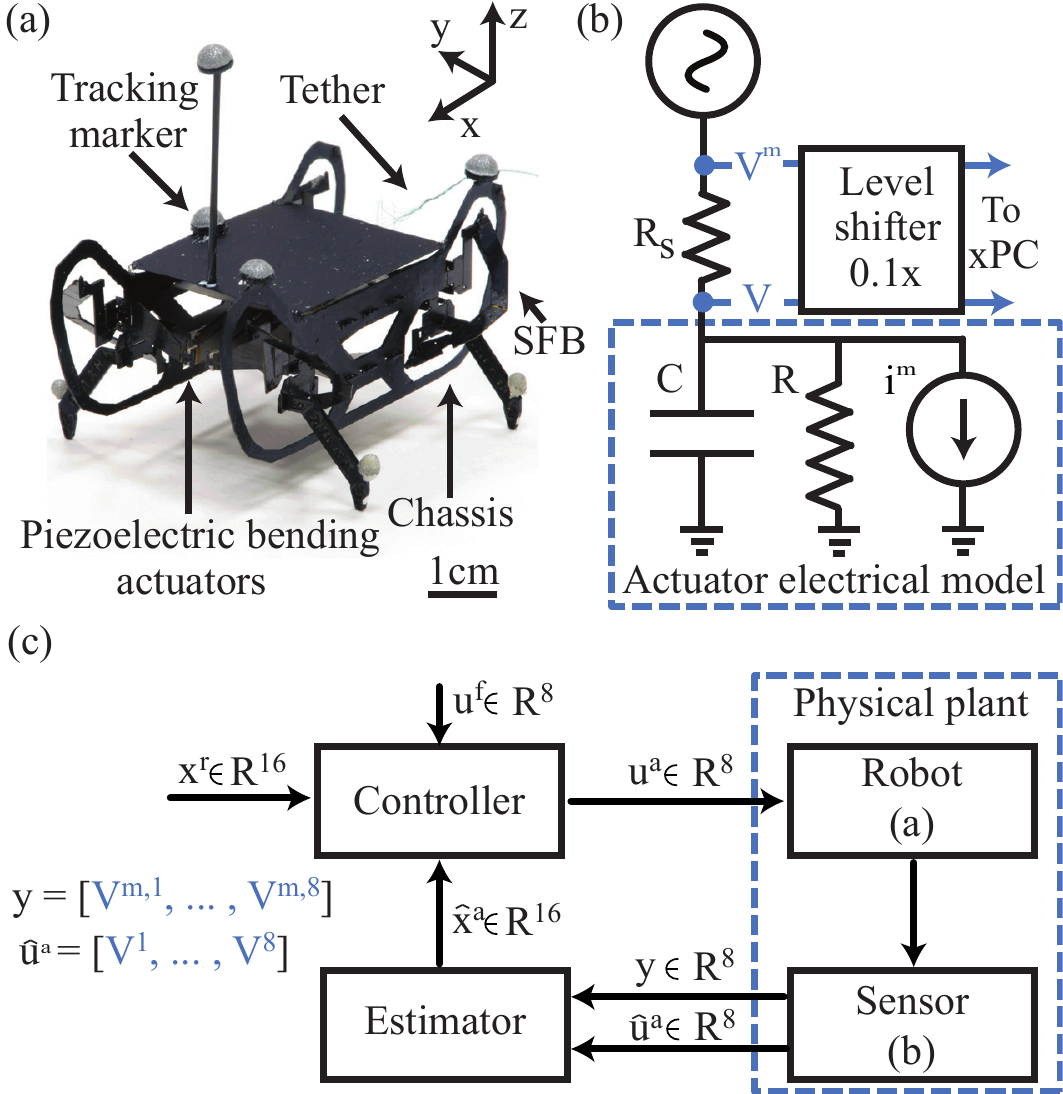}
		\vspace{-0.25cm}
		\caption{(a) Image of HAMR with body-fixed axes shown, and tracking markers and components labeled. (b) Schematic of a lumped parameter electrical model of a single actuator and associated piezoelectric encoder measurement circuit \cite{jayaram2018concomitant}. (c) A block diagram of the proposed sensing and control architecture. Here $x^r$ is the reference actuator position and velocity, $\hat{x}^a$ is estimated actuator position and velocity, $u^f$ is the feed-forward actuator voltage, $u^a$ is the control voltage, and $\hat{u}^a$ and $y$ are the sensor measurements. The design of the estimator and controller are discussed in Secs. \ref{sec:estdesign} and \ref{sec:contdesign}, respectively.  }
		\label{fig:platform_overviewA}
		\vspace{-0.5cm}
	\end{center}
\end{figure}

In contrast, the Harvard Ambulatory MicroRobot (HAMR, Fig.~\ref{fig:platform_overviewA}a) is able to independently control the fore-aft and vertical position of each leg using  high-bandwidth piezoelectric bending actuators. This dexterity enables control over both the shape of individual leg trajectories and gait. Furthermore, HAMR is unique among legged robots in its ability to operate at a wide range of stride frequencies. Despite HAMR's dexterity, however, a lack of sensing and control has limited its operation to using feed-forward sinusoidal voltage inputs resulting in elliptical leg trajectories \cite{baisch2014high, goldberg2017highb}. Though this approach has previously enabled rapid locomotion \cite{goldberg2017high}, high-performance operation (e.g., high speed, low cost-of-transport, etc.) has been limited to a narrow range of stride frequencies \cite{goldberg2017gait}.

In this work, we leverage concomitant sensing for piezoelectric actuation (Fig.~\ref{fig:platform_overviewA}b, \cite{jayaram2018concomitant}) and a computationally inexpensive estimator and controller (Fig.~\ref{fig:platform_overviewA}c) for tracking leg trajectories on a microrobot. The robot and concomitant sensors and discussed in Sec.~\ref{sec:overview}. We then describe the estimator (Sec.~\ref{sec:estdesign}) and controller (Sec.~\ref{sec:contdesign}) and include an important simplification, treating of ground contact as a perturbation. We leverage this capability to track two bioinspired parametric leg trajectories that modulate intra-leg timing, energy, and stiffness (Sec.~\ref{sec:trajdesign}). We experimentally evaluate these trajectories (Sec.~\ref{sec:locomotion_exp}), and demonstrate that our framework enables accurate estimation (Sec.~\ref{sec:est_results}) and tracking (Sec.~\ref{sec:cont_results}) for our operating conditions (\SIrange{10}{50}{\hertz}). Furthermore, we find that these trajectories allow the robot to maintain locomotion performance in the body dynamics frequency regime by reducing leg slip, improving cost-of-transport (COT), and favorably utilizing body dynamics (Sec.~\ref{sec:gait_results}). We generalize these results across the range of operating stride frequencies in Sec.~\ref{sec:gait_summary}, and the discuss implications of this work and potential future extensions in Sec.~\ref{sec:conclusion}.

\section{Platform Overview}
\label{sec:overview}

This section describes the relevant properties of the microrobot (Sec.~\ref{sec:hamr_overview}) and the concomitant sensors (Sec.~\ref{sec:piezo_dyn}).

\subsection{Robot Description}
\label{sec:hamr_overview}

HAMR (Fig.~\ref{fig:platform_overviewA}a) is a \SI{4.5}{\centi\meter} long, \SI{1.43}{\gram} quadrupedal microrobot with eight independently actuated DOFs \cite{doshi2015model}. Each leg has two DOFs that are driven by optimal energy density piezoelectric bending actuators \cite{jafferis2015design}. These actuators are controlled with AC voltage signals using a simultaneous drive configuration described by Karpelson et al. \cite{karpelson2012driving}. A spherical-five-bar (SFB) transmission connects the two actuators to a single leg in a nominally decoupled manner: the swing actuator controls the leg's fore-aft position, and the lift actuator controls the leg's vertical position. A minimal-coordinate representation of the pseudo-rigid body dynamics of this robot has a configuration vector $q = [q^{fb}, q^{a}]^T \in \mathbb{R}^{14}$ and takes the AC voltages signals $u^a \in \mathbb{R}^8$ as inputs. The configuration vector consists of the floating base position and orientation ($q^{fb} \in \mathbb{R}^6$), and the tip deflections of the eight actuators ($q^a \in \mathbb{R}^8$). An alternative minimal-coordinate representation occasionally used in this work is $q^{alt} = [q^{fb}, q^{l}]^T \in \mathbb{R}^{14}$. Here $q^l \in \mathbb{R}^8$ is the vector of the four legs' fore-aft ($l^x$) and vertical ($l^z$) positions, and it is related to $q^a$ by a one-to-one kinematic transformation.

\subsection{Sensor Design and Dynamics}
\label{sec:piezo_dyn}

Eight off-board piezoelectric encoders provide measurements of actuator tip-velocities ($\dot{q}^a \in \mathbb{R}^8$)  \cite{jayaram2018concomitant}. Though these sensors are currently off-board, an on-board implementation is straightforward as the components are both light ($<$ \SI{10}{\milli\gram}) and small ($<$ \SI{5}{\milli\meter}$^2$). Previous work has shown that the tip-velocity of the $i$-th actuator ($\dot{q}^a_i$) is $\alpha$ times the \textit{mechanical current} ($i^{m}$) produced by that actuator's motion; that is, 
\begin{align}
\label{eqn:sensor_dyn1}
\dot{q}^a_i = \alpha i^{m}.
\end{align}
Each encoder (Fig.~\ref{fig:platform_overviewA}b) measures the mechanical current by applying Kirchoff's law to the measurement circuit in series with a lumped-parameter electrical model of an actuator: 
\begin{align}
\label{eqn:sensor_dyn2}
i^{m} = \ \frac{V^{m} - V}{R_s} -  \beta C \dot{V} - \frac{V}{R}. 
\end{align}
The first term on the RHS of Eqn.~\eqref{eqn:sensor_dyn2} is the total current drawn by an actuator computed from measurements of the voltage before ($V^m$) and after ($V$) a shunt resistor ($R_{s}=\SI{75}{\kilo\ohm}$). The actuator is modeled as a capacitor ($C$), resistor ($R$), and current source ($i^m$) in parallel. The voltage and frequency dependent values of $R$ and $C$ have been computed for the range inputs by Jayaram et al.  \cite{jayaram2018concomitant}. Finally,  $\beta$ is a blocking factor which accounts for imperfect measurements of $C$, and is set to 1.57 as described by Jayaram et al. \cite{jayaram2018concomitant}.
 
\section{Estimator Design}
\label{sec:estdesign}

We use the sensors described above in a proprioceptive estimator for leg position and velocity ($x^{a} = [q^{a}, \dot{q}^a]^T \in \mathbb{R}^{16}$). These estimates are used with a feedback controller to command a variety of leg trajectories for improved locomotion. Previous work has focused on the estimation of the floating-base position and velocity for legged systems. This includes approaches that use simplified dynamic models \cite{stephens2011state}, kinematic approaches \cite{bloesch2013state}, hybrid models \cite{singh2007hybrid}, sampling-based techniques (e.g., particle filters \cite{koval2015pose} or unscented Kalman filters \cite{lowrey2014physically}), and more recently, high-fidelity process models that resolve the discontinuous mechanics of ground contact online \cite{varin2018contact}.

For our application, size and payload constraints make it difficult to incorporate additional sensors on the microrobot. This combined with strict computational constraints makes it impractical to use many of the aforementioned approaches. As such, we utilize an infinite-horizon Kalman filter that combines a linear approximation of the transmission model in the absence of contact with the measurement model described in Eqns.~(\ref{eqn:sensor_dyn1}-\ref{eqn:sensor_dyn2}). To simplify the measurement model, we leverage the one-to-one map between leg and actuator position to work in the actuator frame. Our filter averages a drifting position measurement that registers ground contact with a zero-drift position prediction that ignores contact, with the primary advantage that all quantities used in the update rule are pre-computed.

\subsection{Process Model}
\label{sec:proc_model}

Given that we are ignoring ground contact, a single transmission can be modeled in isolation. The minimal-coordinate dynamics of each SFB transmission in the absence of contact is described by the continuous nonlinear difference equation:
\begin{align}
\label{eqn:tranmission model}
{x}_{k+1}^p = f(x_k^p, u_k^p),
\end{align}
where $k$ is the time-step, $x_k^p = [q_k^s, \dot{q}_k^s, q_k^l, \dot{q}_k^l]^T \in \mathbb{R}^4$ is the position and velocity of the swing and lift actuators, and  $u_k^p = [V_k^s, V_k^l]^T \in \mathbb{R}^2$ are the actuator drive voltages. A detailed derivation of $f(x_k^p, u_k^p)$ is presented in Note S1. Instead of calculating the linear approximation of $f(x_k^p, u_k^p)$ about a fixed point $(x^{p}_0, u^p_0)$, we use MATLAB's subspace identification algorithm \texttt{n4sid} \cite{ljung1998system, van1994n4sid} to determine a discrete-time second-order (four-state) linear system that minimizes the prediction error for the range of expected actuator deflections ($\pm$ \SI{0.15}{\milli\meter}) and stride frequencies (\SIrange{10}{50}{\hertz}). While the accuracy of a local linear approximation decreases away from the fixed point, the identified model is accurate in an average sense across the range of expected operating conditions. The resulting discrete-time linear system has the form:
\begin{align}
\label{eqn:process_model}
x^p_{k+1} = A^p x_k^p + B^p (u_k^p - u^p_0) + w_k^p, 
\end{align}
with $A^p \in \mathbb{R}^{4\times4}$ and $B^p \in \mathbb{R}^{4\times2}$. Moreover, the signal $w_k^p \in \mathbb{R}^{4}$ is zero-mean process noise with covariance $W^p$. The \texttt{n4sid} algorithm determines the system matrices ($A^p$ and $B^p$) and noise covariance ($W^p$) of the zero-mean process noise that minimize the squared prediction error in $x^p_k - x^p_0$ when driven with voltages $u^p_k - u^p_0$. We describe the identification process in further detail and evaluate the accuracy of the resulting model in Note S2. Finally, we note that $x^p$ and $u^p$ are subsets of $x^a$ and $u^a$ corresponding to the appropriate transmission, and the identical procedure is carried out to identify a process model for each transmission.

\subsection{Measurement Model}

Since each piezoelectric encoder measurement is independent, the sensor dynamics (Sec.~\ref{sec:piezo_dyn}) is inverted to form the measurement model for a single actuator. We start by combining Eqns.~(\ref{eqn:sensor_dyn1}-\ref{eqn:sensor_dyn2}) with a finite difference approximation of $\dot{V}$ to write a difference equation for $\dot{q}_k$:
\begin{align}
\label{eqn:sensor_discrete}
\dot{q}_k = & c_1 (V^{m}_k - V_k) -c_2 V_k - c_3 (V_k - V_{k-1}), 
\end{align}
where $c_1 = \alpha R_s^{-1}$, $c_2= \alpha R^{-1} $, and $c_3 = \alpha \beta C h^{-1}$. Since Eqn.~\eqref{eqn:sensor_discrete} depends on the previous time-step, we also write a difference equation for $\dot{q}_{k-1}$ using the same finite difference approximation for $\dot{V}_{k-1}$:
\begin{align}
\label{eqn:sensor_discrete2}
\dot{q}_{k-1} = &c_1 (V^{m}_{k-1} - V_{k-1} ) -c_2 V_{k-1} -c_3 (V_k - V_{k-1}).
\end{align}

Combining Eqns.~(\ref{eqn:sensor_discrete}) and (\ref{eqn:sensor_discrete2}) and solving for $y^m_k = [V^{m}_k, V^{m}_{k-1}] \in  \mathbb{R}^{2}$ gives the measurement model: 
\begin{align}
\label{measurment}
y^m_k = H^m x^m_k + D^m u^m_k + n^m_k.
\end{align}
Here
\begin{align}
\label{eqn:measurment_model}
H^m &=  \frac{1}{c_1}  \begin{bmatrix} 0_{2 \times 1} &  I_{2 \times 2} \end{bmatrix} \in \mathbb{R}^{2\times3},\\
D^m &=  \frac{1}{c_1} \begin{bmatrix}c_1 + c_2 + c_3 & -c_3 \cr c_3 & c_1 + c_2 + c_3 \end{bmatrix} \in \mathbb{R}^{2\times2},     
\end{align} 
$x^m_k$ =  $[q_k, \dot{q}_k, \dot{q}_{k-1}]^T \in \mathbb{R}^{3}$, and $u^m_k = [V_k, V_{k-1}] \in \mathbb{R}^{2}$. The signal $n^m_k \in \mathbb{R}^2$ is zero-mean measurement noise with covariance $ N^m = N^{H} + D N^{D} D^T$. The measurement noise covariance is computed directly on the hardware, and we describe our process in Sec.~\ref{sec:calibration}. Note that the process and measurement states are not equal ($x^m \neq x^p$), and the following section builds an augmented state to resolve this discrepancy. 

\subsection{Complete Estimator}
Combining the process and measurement models, we write the linearized discrete-time dynamics of a single transmission-sensor system in the following form: 
\begin{align}
\label{eqn:lin_mod}
x_{k+1} =& A x_k + B u_k + w_k  \\
y_k =& H x_k + D u_k +  n_k, \
\end{align}
where $x_k = [(x^p_k)^T, \dot{q}^s_{k-1}, \dot{q}^l_{k-1}]^T \in \mathbb{R}^{6}$ is the state, $y_k = [V^{m,s}_k, V^{m,l}_k, V^{m,s}_{k-1}, V^{m,l}_{k-1}]^T \in \mathbb{R}^{4}$ is the measurement, $u_k = [(u^p_k)^T, (u^p_{k-1})^T]^T \in \mathbb{R}^{4}$ is the input. Furthermore, $w_k \in \mathbb{R}^{6}$ and $n_k \in \mathbb{R}^{4}$ are the zero-mean process and measurement noise with covariance given by
\begin{align}
W = \begin{bmatrix} W^p & 0 \cr 0 & 0 \end{bmatrix} \quad \text{and} \quad  N = \begin{bmatrix}N^m & 0 \cr 0 & N^m \end{bmatrix}, 
\end{align}
respectively. Finally, the system matrices are given by 
\begin{align}
A &= \begin{bmatrix} A^p  & 0_{4 \times 2} \cr [e_2, e_3]^T & 0_{2 \times 2} \end{bmatrix} \in \mathbb{R}^{6\times6}, \\
B &= \begin{bmatrix} B^p & 0_{2 \times 2} \cr 0_{2 \times 2}  & 0_{2 \times 2} \end{bmatrix} \in {R}^{6\times4}, \\ 
H &= \begin{bmatrix} 0 & h^m_{11} & 0_{1 \times 4} \cr 0_{1 \times 2} & h^m_{11} & 0_{1 \times 3} \cr 0_{1 \times 4} & h^m_{22} & 0  \cr 0_{1 \times 4} & 0 & h^m_{22} \end{bmatrix} \in \mathbb{R}^{4 \times 6}, \\
D &= \begin{bmatrix}d^m_{11} & 0 & d^m_{12} & 0 \cr 0 & d^m_{11} & 0 & d^m_{12} \cr  d^m_{21} & 0 & d^m_{22} & 0 \cr 0 & d^m_{21} & 0 & d^m_{22} \end{bmatrix} \in \mathbb{R}^{4 \times 4}.
\end{align}
Here $h^m_{ij}$ and $d^m_{ij}$ are $ij$-th entries of $H^m$ and $D^m$, respectively, and $e_2$ and $e_3$ are elementary unit vectors in $\mathbb{R}^4$. Given this formulation, the infinite-horizon Kalman gain is computed off-line as $K = P H( H P H^T + R)^{-1} \in \mathbb{R}^{6 \times 4}$, where $P$ is found by solving the discrete-time algebraic Ricatti equation \cite{arnold1984generalized}. The current state estimate is then given by
\begin{align}
\label{eqn:kalman_estimate}
\hat{x}_{k} = & A \hat{x}_{k-1} + B u_{k-1} + \\ \nonumber 
& K \big(y_k - H(A\hat{x}_{k-1} + B u_{k-1}) - D u_k\big),
\end{align}
where the state is initialized to $\hat{x}_0 = 0$. 

This simple update rule can be carried out independently for each transmission and only requires the addition of vectors $\mathbb{R}^6$ and multiplication of vectors in $\mathbb{R}^6$ by sparse matrices in $\mathbb{R}^{6\times6}$. Though this filter is currently implemented off-board, this method, because of its computational efficiency, can easily be implemented in real-time on the autonomous version of this robot \cite{goldberg2018power}.

\section{Controller Design}
\label{sec:contdesign}

Similar to the complete estimator, the feedback controller is also independently derived for a transmission-sensor system. A subset of estimated actuator positions and velocities ($\hat{x}_k^p$) is used in a feedback controller designed as a linear-quadratic-regulator (LQR). LQR controllers have been used to stabilize both smooth and hybrid non-linear systems; for example, the time-varying LQR formulation (TVLQR, \cite{kwakernaak1972linear}) is often used to locally stabilize nonlinear systems about a given trajectory. Furthermore, LQR has been used to stabilize limit cycles for hybrid systems, both in full-coordinates using the jump-Ricatti equation \cite{dai2012optimizing} and in transverse-coordinates using a transverse linearization \cite{manchester2011stable}.

In this work, since each of HAMR's leg can exert forces greater than one body-weight \cite{doshi2015model}, we can treat the relatively small contact forces as disturbances. Furthermore, since an LTI system provides an accurate representation of the transmission dynamics in air, we choose to use an infinite-horizon LQR controller. This controller minimizes the following cost function: 
\begin{align}
J = \sum_{k=0}^{\infty} (\hat{x}^p_k - x^0_k)^T Q &(\hat{x}^p_k - x^0_k) + \nonumber \\ &(u^p_k - u^p_0)^T R (u^p_k - u^p_0),  
\end{align}
where $Q \succeq 0$ and $R \succ 0$ are symmetric matrices that penalize deviations from the fixed point $(x^{p}_0, u^p_0)$. We defined $Q$ and $R$ as diagonal matrices parameterized by three positive scalars ($k_p$, $k_d$, and $k_u$) that determine trade-offs between squared deviations in actuator position, velocity, and control voltage, respectively. The complete control law combines the LQR feedback rule with a feed-forward term ($u^f_k = u^p_0 + u^{t}_k \in \mathbb{R}^2$): 
\begin{align}
\label{eqn:controller}
u^p_{k} = u^f_k + L(x^r_k - \hat{x}^p_k).
\end{align}
Here $x^r_k \in \mathbb{R}^4$ is the reference state, $L=(R+B^TSB)^{-1} B^T S A \in \mathbb{R}^{2\times4}$ is the feedback matrix, and $S$ is computed by solving the discrete-time algebraic Ricatti equation \cite{arnold1984generalized}. The resulting linear-quadratic-Gaussian (LQG) dynamical system is formulated by combining Eqn.~\eqref{eqn:kalman_estimate} with the control law given in Eqn.~\eqref{eqn:controller}. 

Intuitively, the feed-forward term is equal to the nominal voltage ($u^p_0$) if the reference state is the fixed point. Furthermore, the control law in Eqn.~\eqref{eqn:controller} will stabilize the LQG system since $Q$ and $R$ are chosen to be positive-definite. In practice, the controller is used to track reference trajectories on the physical (nonlinear) legged robot, the control input ($u^p_k$) still acts to reduce the error, and ground reaction forces can be thought of as disturbances. We also augment the feed-forward term with a time varying component ($u^t_k$) that is computed via a trajectory optimization without ground contact (Note S3). This term is similar to the nominal input for a TVLQR controller about a trajectory; however, the lack of ground contact modeling makes it more of a heuristic for improving the convergence rate and reducing steady-state error. 

\section{Bio-inspired Trajectory Selection}
\label{sec:trajdesign}

Using the estimation and control framework described in the previous two sections (Secs.~\ref{sec:estdesign} and \ref{sec:contdesign}), we are now able to track arbitrary leg trajectories subject to the dynamics of the transmission. We exploit this to expand on our previous work that explored the effect of gait and stride frequency on locomotion \cite{goldberg2017gait}. The major challenges that limited locomotion performance in our previous studies are: 
\begin{enumerate}[(1)]
    \item  High leg-slip (\SIrange{40}{45}{}\% ineffective stance) across all stride frequencies. 
    \item  Increased body oscillations (in roll and pitch) in the body dynamics frequency range (\SIrange{20}{40}{\hertz}).
    \item Departure from SLIP-dynamics \cite{cavagna1977mechanical} beyond the mechanically tuned operating point close to robot $z$-resonance ($\sim$\SI{10}{\hertz}). 
    \item  Fixed (open-loop) timing between vertical and fore-aft resulting in poor or backwards locomotion (e.g., when pronking at \SI{10}{\hertz}).
\end{enumerate}

\begin{figure*}[t]
	\begin{center}
		\includegraphics[width=\textwidth]{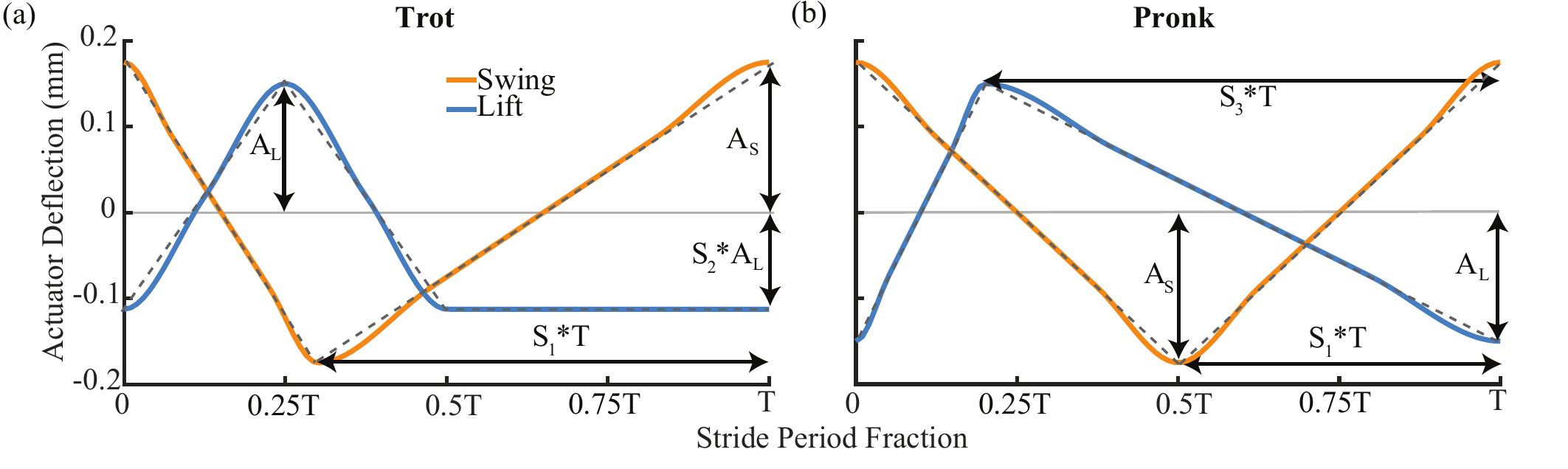}
		\vspace{-0.25cm}
		\caption{(a) Reference actuator positions for the swing (orange) and lift (blue) for the trot gait leg trajectory with $S_1 = 70 $, and $S_2 = 75 $. (b) The same for the pronk gait leg trajectory with $S_1 = 50 $, and $S_3 =80 $. Note that $A_S$ and $A_L$ are fixed to the values given in Tab.~\ref{tab:trajdesign}, and the smooth reference trajectories (in orange and blue) are generated by fitting a cubic-spline to the non-smooth desired trajectories (grey dashed lines). }
		\label{fig:traj_design}
		\vspace{-0.5cm}
	\end{center}
\end{figure*}

\begin{table*}[t]
\centering
\caption{Heuristic trajectory design parameters}
\vspace{0.1cm}
\label{tab:trajdesign}
\begin{tabular}{ >{\centering\arraybackslash}m{2cm} |  >{\centering\arraybackslash}m{2.5cm} | 
>{\centering\arraybackslash}m{5.0cm} | 
>{\centering\arraybackslash}m{5.0cm} }

\textbf{Parameters} & \textbf{Description} & \textbf{Trot Gait} & \textbf{Pronk Gait} \\ \hline
$A_S$ & swing amplitude& \SI{175}{\micro\meter} & \SI{150}{\micro\meter} \\ \hline
$A_L$ & lift amplitude & \SI{175}{\micro\meter} & \SI{150}{\micro\meter} \\ \hline
$T$ & stride period (\nicefrac{1}{\text{frequency}}) & $\in [\nicefrac{1}{50}, \nicefrac{1}{40}, \nicefrac{1}{30}, \nicefrac{1}{20}, \nicefrac{1}{10}]$ \SI{}{\milli\second} &$\in [\nicefrac{1}{50}, \nicefrac{1}{40}, \nicefrac{1}{30}, \nicefrac{1}{20}, \nicefrac{1}{10}]$ \\ \hline
$S_1$ & shape control one & leg retraction period (\%$T$) $\in [50, 60, 70, 80]$ & leg retraction period (\%$T$) $\in [50, 60, 70, 80]$\\ \hline
$S_2$ & shape control two & maximum leg adduction (\%$A_l$) $\in [-75, -50, -25, 0, 25]$  & N/A \\ \hline
$S_3$ & blueshape control three & N/A & leg adduction period (\%$T$) $\in [20, 35, 50, 65, 80]$\\ 
\end{tabular}
\end{table*}

In this work, we postulate the following four specific hypotheses to understand the underlying mechanisms behind the challenges enumerated above. These hypotheses (described below) are motivated by relevant examples from recent scientific literature, and the application of these ideas to an dexterous insect-scale system across a wide range of stride frequencies is a contribution of this work. Ultimately, we hypothesize ($H_0$) that exploring the leg trajectories described below can reveal optimized shape control parameters that enable high-performance locomotion over the entire operating range of the robot, overcoming challenges observed in our previous research \cite{goldberg2017gait}.

\subsection{Hypothesis One ($H_1$)}
Template models of legged locomotion, such as SLIP, have relied on a swing-leg retraction strategy for stabilizing sagittal plane locomotion \cite{seyfarth2003swing, wisse2005swing, hobbelen2008swing, vejdani2013bio, piovan2015reachability}. These results have been supported by numerous experimental studies on bipedal running \cite{muller2016human} in humans \cite{seyfarth2003swing, blum2010swing} and guinea fowls \cite{daley2006running, daley2018understanding}, and on quadrupedal galloping in  horses \cite{herr2001galloping}. Expanding this approach, researchers have demonstrated an optimal retraction rate for perturbation rejection \cite{karssen2011optimal} and energy efficient locomotion \cite{haberland2011effect}. Additionally, modeling and experimental results using large bio-inspired quadrupedal robots \cite{karssen2011optimal, karssen2015effects} indicate that swing leg retraction can potentially mitigate the risk of slippage at heel-strike during rapid running. Therefore, we test the effect of varying leg retraction period on locomotion and hypothesize ($H_1$) that \textit{increasing the leg retraction period reduces slipping and improves locomotion performance.} 

\subsection{Hypothesis Two ($H_2$)}
Upright-posture animals have been shown to modulate their normal force and vertical impulse to minimize body oscillations and maintain stable locomotion in the sagittal plane \cite{pearson1976control, pearson2004generating, ting2005limited, chvatal2013common}. Similarly, studies in humans show that the above considerations are important for overcoming roll perturbations and achieving lateral stability \cite{hof2010balance, arellano2011effects}. Robots employ these bio-inspired strategies \cite{raibert1986legged, raibert1986running, srinivasan2008well, qiao2012task} to stabilize hip height \cite{hodgins1991adjusting, hodgins1996three} and control pitch oscillations \cite{pratt2000exploiting, kim2007walking}. The underlying mechanisms either passively (mechanically) \cite{gregorio1997design, poulakakis2006stability} or actively modulate ground reaction forces \cite{koepl2010force} and impulses \cite{koepl2014impulse, park2015quadrupedal}. We adapt this approach to minimize vertical, pitch, and roll body oscillations in the body dynamics frequency range, and we hypothesis ($H_2$) that \textit{increasing input lift energy, especially in the body dynamics frequency range, increases detrimental body oscillations and reduces locomotion performance.}

\subsection{Hypothesis Three ($H_3$)}
Animals of varying size and morphology \cite{blickhan1993similarity} use energy storage and exchange mechanisms \cite{mcmahon1985role, kram1990energetics, mcmahon1990mechanics} during locomotion \cite{dickinson2000animals}. Numerous models explain these ubiquitous underlying mechanisms, the most popular of which is the SLIP model \cite{cavagna1977mechanical, blickhan1989spring, holmes2006dynamics, seyfarth2002movement}. Furthermore, the implications of relative stiffness \cite{heglund1974scaling, blickhan1993similarity} on locomotion speed \cite{farley1993running, farley1996leg, weyand2009fastest, mcmahon1990mechanics}, stability \cite{ferris1998running, schmitt2002dynamics} and economy \cite{taylor1970scaling, snyder2011energetically, kram1990energetics, kim2011leg} are well documented across body sizes. Based on this understanding, we hypothesize ($H_3$) that \textit{increasing effective leg stiffness allows for greater energy storage and return (SLIP-like dynamics) and improves performance at higher stride frequencies.}

\subsection{Hypothesis Four ($H_4$)}
During running the body decelerates during the first half of stance, and accelerates into flight during the second half of the stance. Studies have shown that relative timing of vertical and fore-aft leg motions is important in achieving a pattern of deceleration and acceleration that results in effective locomotion \cite{hamner2013muscle, hasaneini2013optimal}. Given that time-of-flight will change as a function of stride frequency (due to body resonances), we hypothesize ($H_4$) that \textit{the timing between the vertical and fore-aft leg motions that results in the best performance varies as a function of stride frequency.}

\subsection{Trajectory Design}

We distill these four hypotheses into parametric leg trajectories for the trot (Fig.~\ref{fig:traj_design}a) and pronk (Fig.~\ref{fig:traj_design}b, supplementary video S4) gait, respectively. Each trajectory is defined by five parameters described in Tab.~\ref{tab:trajdesign}. Here, the swing ($A_S$) and lift ($A_L$) actuator amplitudes are held constant, $T$ controls the stride frequency, and the shape parameters $S_1$, $S_2$, and $S_3$ vary as described below. For both parametric trajectories, we address $H_1$ by maintaining a constant speed during leg retraction and vary the leg retraction period as a trajectory shape control parameter $S_1$. For the trot gait, we also vary the maximum leg adduction via the shape parameter $S_2$. This modification directly varies the net energy imparted to the lift ($z$) motion addressing $H_2$. In addition $S_2$ also modulates leg stiffness (see Fig S2 and Note S3) addressing $H_3$. Finally, we vary the leg adduction period as the trajectory shape control parameter $S_3$ for the pronk gait. This modification, coupled with $S_1$ from above, varies the timing between the vertical and fore-aft leg motions addressing $H_4$.\\

\section{Experimental Design, Methods and Metrics}
\label{sec:locomotion_exp} 

This section first describes the calibration conducted before running experiments (Sec~\ref{sec:calibration}). We then describe the experimental procedures and apparatus for evaluating the estimator (derived in Sec.~\ref{sec:estdesign}) and controller (derived in Sec.~\ref{sec:contdesign}) performance, and for exploring the heuristic leg trajectories (developed in Sec.~\ref{sec:trajdesign}).  Finally, we define a number of locomotion performance metrics in Sec.~\ref{sec:perf_metrics} that are used to quantify the effects of varying leg trajectory shape in Sec.~\ref{sec:gait_results}. 

\begin{figure*}[t]
	\begin{center}
		\includegraphics[width=\textwidth]{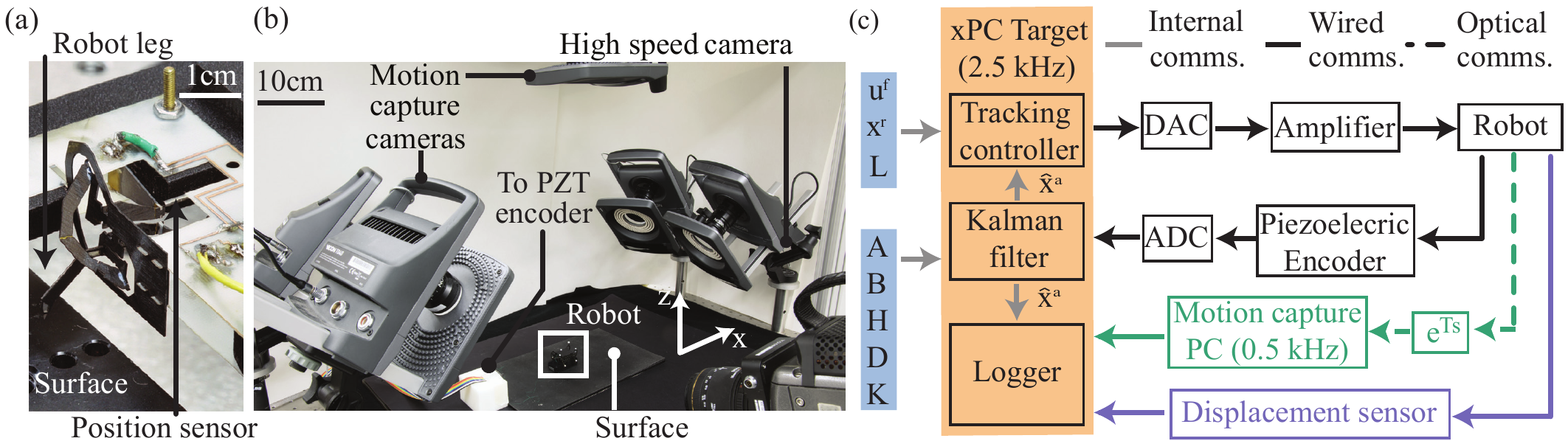}
		\vspace{-0.5cm}
		\caption{(a) Experimental setup with single-leg used to evaluate estimator performance with components labeled. Ground truth is provided by a calibrated fiber-optic displacement sensor (Philtec-D21) at \SI{2.5}{\kilo\hertz}. (b) Perspective image of the locomotion arena used to evaluate the controller and explore the heuristic locomotion trajectories. Important components and world-fixed axes are labeled. (c) Augmented communication and control block diagram for the experimental setups shown in (a) and (b). The displacement sensor (purple) is used as ground-truth in (a), and a motion capture system (Vicon, T040; green) is used as ground-truth in (b). The Kalman filter and tracking controller run on the xPC target (shaded in orange) at \SI{2.5}{\kilo\hertz}. Reference actuator trajectories ($x^r$), the feedback control-law ($L$), the Kalman update (matrices $A$, $B$, $H$, $D$, and $K$), and the feed-forward control signals ($u^{ff}$) are shaded in blue and are pre-computed off-line.}
		\label{fig:exp_setup1}
		\vspace{-0.cm}
	\end{center}
\end{figure*}

\subsection{Calibration}
\label{sec:calibration}

A calibration was performed for each robot and single-leg before conducting all experiments. The measurement noise covariances $N^{H}$ and $N^{D}$ were computed from mean-subtracted measurements of $V^{m}$ and $V$, respectively, with $u^a = 0$. These means (corresponding to an initial offset) were also subtracted from subsequent measurements of $V^{m}$ and $V$. The velocity scaling coefficients ($\alpha$) from the mechanical current (\SI{}{\milli\ampere}) to tip velocity (\SI{}{\milli\meter\second^{-1}}) were computed for each actuator over the range of operating frequencies. The coefficient for each actuator was set to the value that minimized the squared-error between the mechanical current ($i_{m}$, Eqn.~\eqref{eqn:sensor_dyn2}), and corresponding ground-truth leg velocity.

\subsection{Estimator Validation}
\label{sec:est_valid_exp}

Estimator validation was conducted on a single-leg (Fig.~\ref{fig:exp_setup1}a) using the architecture shown in Fig \ref{fig:exp_setup1}c. Note that control gains ($L$) were set to zero. Sinusoidal input signals ($u^f$) were generated at \SI{2.5}{\kilo\hertz} using a MATLAB xPC environment (MathWorks, R2015a), and were supplied to the single-leg through a four-wire tether. The Kalman filter (defined in Sec.~\ref{sec:estdesign}) estimated actuator position and velocity from the voltage measurements provided by two piezoelectric encoders at \SI{2.5}{\kilo\hertz}. Finally, ground truth swing and lift actuator position measurements were provided by calibrated fiber-optic displacement sensors (Philtec-D21) at the same rate. 

We measured estimator performance at stride frequencies of 10, 20, 30, 40, and \SI{50}{\hertz} both in-air and with ground-contact. Ground contact was achieved by positioning a surface at the neutral position of the leg for the duration of the trial. Estimation error for a single actuator was quantified as $\bar{E}_{est}$, which is the $N$-cycle mean of the RMS error between the estimated actuator position and ground-truth measurements normalized by the peak-to-peak amplitude of the ground-truth measurements. 

We also quantified estimator performance on a full-robot at frequencies of 10, 20, 30, 40, and \SI{50}{\hertz} using the locomotion arena shown in blue Fig.~\ref{fig:exp_setup1}b to determine if the estimator could also be used to accurately predict leg positions $(l^x, l^z)$.These trials were also conducted using the architecture shown in Fig.~\ref{fig:exp_setup1}c with sinusoidal inputs and the control gains set to zero. Five motion capture cameras (Vicon T040) tracked the position and orientation of the robot at \SI{500}{\hertz} with a latency of \SI{11}{\milli\second}. A custom C++ script using the Vicon SDK enabled tracking of the leg tips in the body-fixed frame. We used a model of the transmission kinematics to map the estimated actuator position to $l^x$ and $l^z$, and these estimates were compared against ground truth leg position measurements provided by the motion-capture system. Performance was quantified using $\bar{E}_{est}$. 

\subsection{Controller Validation}

We also quantified controller performance on an entire robot at frequencies of 10, 20, 30, 40, and \SI{50}{\hertz}. Experiments were performed both in air and in the presence of ground contact using the experimental arena shown in Fig.~\ref{fig:exp_setup1}c and described in Sec.~\ref{sec:est_valid_exp}. To determine the effectiveness of the controller performance, we quantified tracking error using $\bar{E}_{cont}$, defined as the $N$-cycle mean of the RMS error between the estimated and desired actuator position measurements normalized by the peak-to-peak amplitude of the desired actuator position.

\subsection{Leg Trajectory Exploration}
\label{sec:closedlooplegexploration}

Finally, we performed 400 closed-loop trials to evaluate HAMR's performance when using the two classes of heuristic leg trajectories (Sec.~\ref{sec:trajdesign}). These experiments used two robots whose floating-base natural frequencies are characterized in Fig.~S1 using methods described by Goldberg et al. \cite{goldberg2017gait}. Two hundred trials were conducted on each robot with 100 trials for each class of heuristic leg trajectory. Each subset of one hundred trials enumerated all possible combinations of stride period ($T$) and shape parameters ($S_1$, $S_2$, and $S_3$). The 400 trials were all conducted in the locomotion arena described above. Since both robots showed similar performance, we averaged the data to compute locomotion metrics (Sec.~\ref{sec:perf_metrics}). 

\subsection{Locomotion Performance Metrics}
\label{sec:perf_metrics}

\subsubsection{Normalized per Cycle Speed ($\nu$)}
\label{sec:normalize_speed}
This is a measure of the speed of the robot ($v$) during locomotion. It is the defined as the ratio of speed achieved per step to the kinematic step length and is computed as
\begin{align}
\label{eqn:norm_cyc_speed}
\nu =  \frac{v}{L_snf},
\end{align}
where $L_s =$ \SI{4.7}{\milli\meter} is the kinematic step length, $n$ is the number of steps per stride for a given gait ($n_{trot}=2, n_{pronk}=1$) and $f =\frac{1}{T}$ is the stride frequency. Intuitively, $\nu=1$ is the expected forward speed assuming ideal kinematic locomotion, and $\nu>1$ suggests that the robot is utilizing dynamics favorably to increase its stride length beyond the kinematic limits. 

\subsubsection{Step Effectiveness ($\sigma$)}
\label{sec:step_effectiveness}
This is a measure of the robot leg slippage during locomotion. It is defined for each leg as one minus the ratio of leg-slip to the kinematic step length. We consider leg-slip to be the total distance a single leg travels in the direction opposite to the robot heading in the world frame. We present an average value for all four legs computed as 
\begin{align}
\label{eqn:norm_step_effectiveness}
\sigma = 1-\frac{1}{4L_s}\sum_{i=1}^{4} \int_{\zeta}|v_x^i(t)|dt,
\end{align}
where $v_x^i$ is the x-velocity of the $i^{th}$ leg in the world-fixed frame, and
$\zeta$ is the set of times within a step for which $v_x^i$ is in the opposite direction as the robot heading. Intuitively, $\sigma=1$ indicates no slipping while $\sigma=0$ indicates continuous slipping (i.e., no locomotion of the robot).

\subsubsection{Locomotion Economy ($\epsilon$)}
\label{sec:locomotion economy}
This is a measure of the the robot's COT \cite{alexander2003principles}. This is defined as the ratio of the robot's mechanical output power to the total electrical power consume and is quantified as: 
\begin{align}
\label{eqn:locomotion_economy}
\epsilon = \frac{mgv_x}{\sum_{i=1}^{8}\frac{1}{T} \int_{0}^{T}i^m(t)V^m(t) dt},
\end{align}
where $m$ = \SI{1.43}{\gram} is the mass of the robot and $g$ = \SI{9.81}{\meter\second^{-2}}) is the acceleration due to gravity. Intuitively, lower values of $\epsilon$ indicate poor conversion of the input electrical power into mechanical output, suggesting ineffective locomotion performance.  

\subsection{Open-loop Control Trajectory Comparison}

\subsubsection{Coupled Sinusoids}
\label{sec:coupledmatching}
 The RMS amplitude for each sinusoidal drive voltage was equal to the average of the RMS voltages delivered to all eight actuators during the fastest trial at a particular stride period. 
This control experiment did not discriminate between voltages delivered to the lift and swing DOFs and is therefore referred to as the \textit{coupled configuration}.  

 \subsubsection{Decoupled Sinusoids}
 \label{sec:uncoupledmatching}
The RMS amplitude for the four lift (and four swing) actuators was equal to the average RMS voltage delivered to the lift (and swing) actuators during the fastest trial at a particular stride period, respectively. The voltages delivered to the lift and swing actuators were individually computed, and therefore, this is referred to as the \textit{decoupled configuration.}

\section{Estimator and Controller Performance}

This section summarizes our results related to the quantification of estimator and controller performances. In particular, we evaluate the accuracy of both the linear approximation (described in Sec.~\ref{sec:proc_model}) of the transmission model and the treatment of ground contact as a perturbation (described in Sec.~\ref{sec:contdesign}).

\subsection{Estimator}
\label{sec:est_results}

\begin{figure}[t]
	\begin{center}
		\includegraphics[width=\columnwidth]{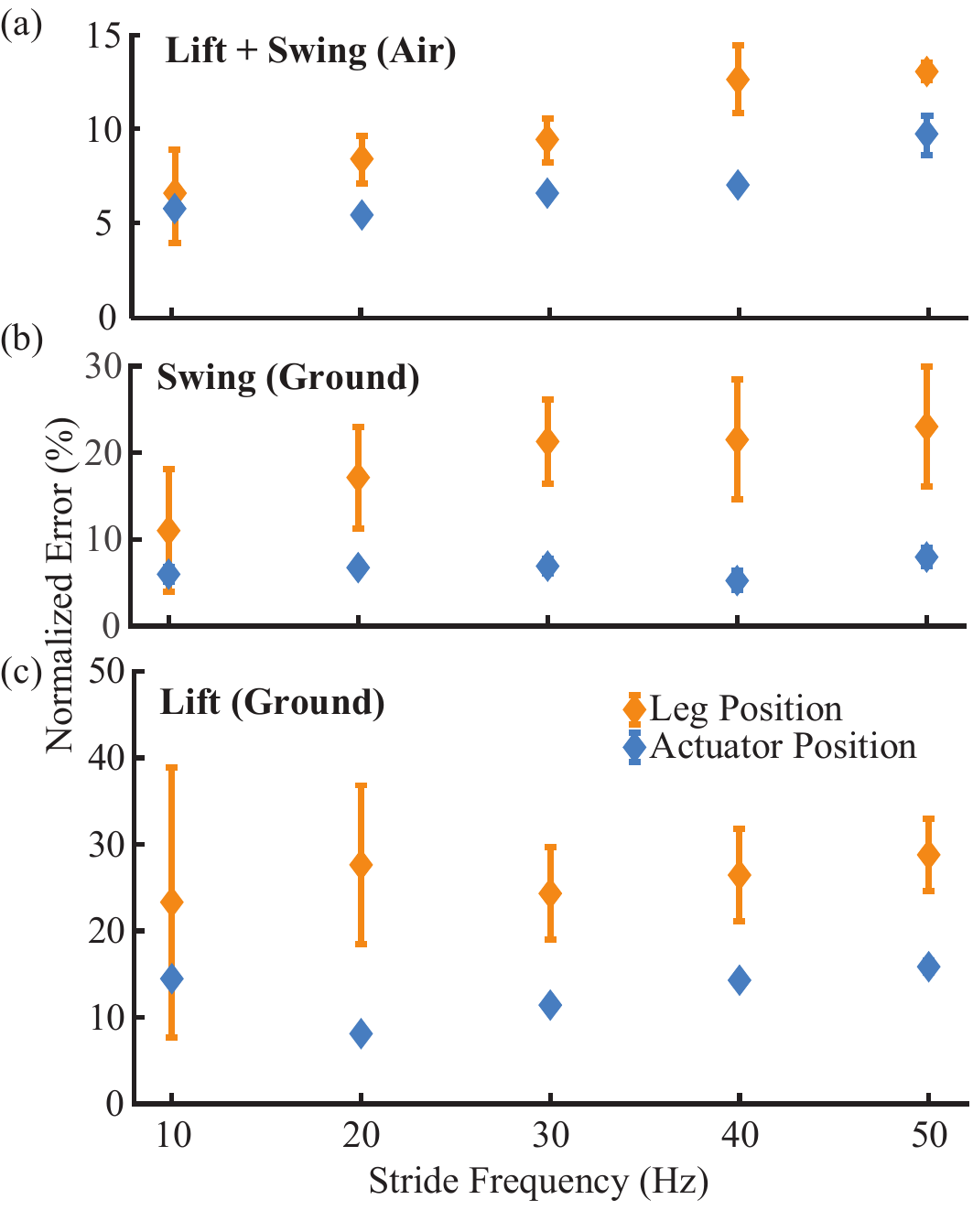}
		\vspace{-0.25cm}
		\caption{(a) Mean and standard deviation in normalized estimator error ($\bar{E}_{est}$) without ground contact in actuator position (blue, one transmission, one robot) and leg position (orange, four transmissions, one robots) as a function of stride frequency. (b) Mean and standard deviation for normalized estimator error with ground contact in actuator position (blue, one transmission, one robot) and leg position (orange, eight transmissions, two robot) for the swing DOF (top) and the lift DOF (bottom). All values of normalized estimation error for (a) and (b) are computed across 15 cycles. } 
		\label{fig:estimator_performance}
		\vspace{-0.5cm}
	\end{center}
\end{figure}

The performance of the estimator is shown in Fig.~\ref{fig:estimator_performance} with estimation errors for a representative trial in air and on the ground shown in supplementary Fig.~S4. For the trials in air (Fig.~\ref{fig:estimator_performance}a), the mean normalized estimation error in actuator position ranges from $5\%$ at \SI{10}{\hertz} to $10\%$ at \SI{50}{\hertz}. These numbers indicate reasonably accurate estimation in air, confirming the accuracy of the sensor measurements and validity of the linear approximation of the non-linear transmission dynamics. The error in leg position is higher than actuator position error and ranges from $6$\% at \SI{10}{\hertz} to $15\%$ at \SI{50}{\hertz}, and we suspect this is due to inaccuracies in the modeled transmission kinematics. 

Similarly, actuator position error (blue) is low when subject to approximated ground-contact. The normalized swing (Fig.~\ref{fig:estimator_performance}b) and lift (Fig.~\ref{fig:estimator_performance}c) actuator position errors are between $5\%$-$10\%$ and $8\%$-$16\%$, respectively. We suspect that the lift position errors are higher because the process model does not capture the effect of (1) perturbations from ground contact and (2) serial compliance between the actuator and mechanical ground \cite{ozcan2014powertrain}. Nevertheless, these errors are still relatively small, indicating that the Kalman filter effectively averages the sensor measurement that registers contact with a linear predication that does not drift.

Finally, we find that the normalized leg position error (orange) with ground contact is higher than normalized actuator position-error (blue). The leg-$x$ error ($l_x$, Fig.~\ref{fig:estimator_performance}b) ranges from $11\%$ to $24\%$, and leg-$z$ error ranges ($l_z$, Fig.~\ref{fig:estimator_performance}c) from $23\%$ to $29\%$. The most likely cause of this is the serial compliance in the transmission---a common problem in flexure-based devices \cite{doshi2015model, ozcan2014powertrain}. This serial compliance alters the kinematics of the transmission by effectively adding un-modeled DOFs between the actuators and leg and changes the assumed one-to-one mapping between actuator and leg positions.

\subsection{Controller}
\label{sec:cont_results}

\begin{figure}[ht]
	\begin{center}
		\includegraphics[width=\columnwidth]{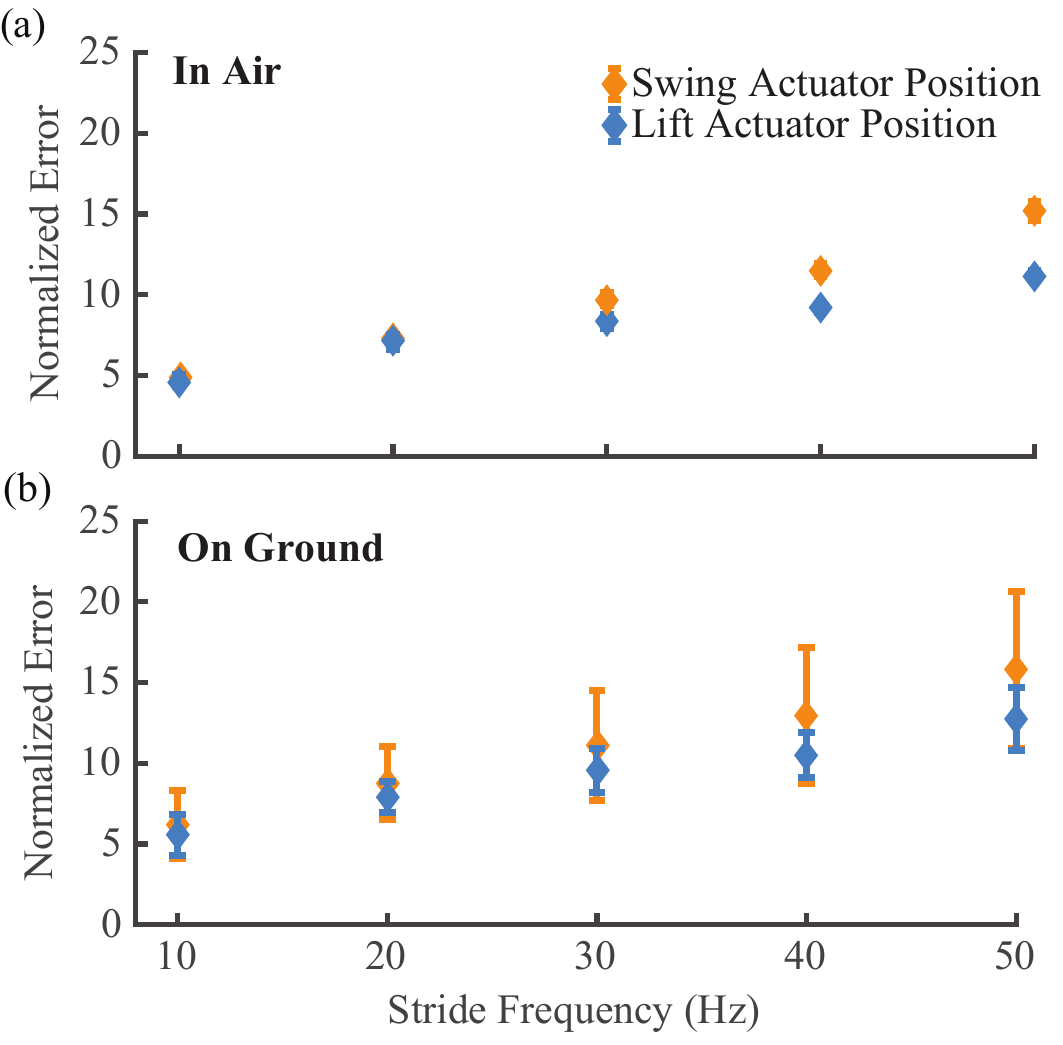}
		\vspace{-0.25cm}
		\caption{Normalized tracking error ($\bar{E}_{cont}$) in swing (orange) and lift (blue) actuator positions as a function of frequency. (a) Normalized tracking error in air. The mean and standard deviation at each frequency is computed using the same robot across four transmissions and $n=60$ cycles. (b) Normalized tracking error when running on a card-stock surface. The mean and standard deviation at each frequency is computed using two different robots across eight transmissions and $n=1200$ cycles. Note that mean normalized tracking error when running on the ground is approximately equal to the same in air.}
		\label{fig:cont_performance}
		\vspace{-0.5cm}
	\end{center}
\end{figure}

The performance of the controller is shown in Fig.~\ref{fig:cont_performance} with tracking errors for a representative trial in air and on the ground shown in supplementary Fig.~S5. For the trials in air (Fig.~\ref{fig:estimator_performance}a), the mean normalized estimation error in actuator position increases from $5\%$ at \SI{10}{\hertz} to $15\%$ at \SI{50}{\hertz} for the swing DOF and from $5\%$ at \SI{10}{\hertz} to $11\%$ at \SI{50}{\hertz} for the lift DOF. This demonstrates the linear approximation of the transmission dynamics is sufficient for control in the absence of ground contact. Moreover, the normalized tracking error (Fig.~\ref{fig:cont_performance}b) for both the swing and lift DOFs when running is also small, and it increases from $6\%$ at \SI{10}{\hertz} to $16\%$ at \SI{50}{\hertz}. This indicates that treating ground contact as a perturbation does not significantly reduce tracking performance. Finally, a likely reason for the increase in tracking error as a function of stride frequency is that the high-frequency components in the heuristically designed leg trajectories become harder to track as they approach the robot's transmission resonant frequencies (between \SIrange{80}{100}{\hertz}, \cite{doshi2017phase}).

\section{Locomotion performance}
\label{sec:gait_results}

\begin{figure*}[ht]
	\begin{center}
		\includegraphics[width=2\columnwidth]{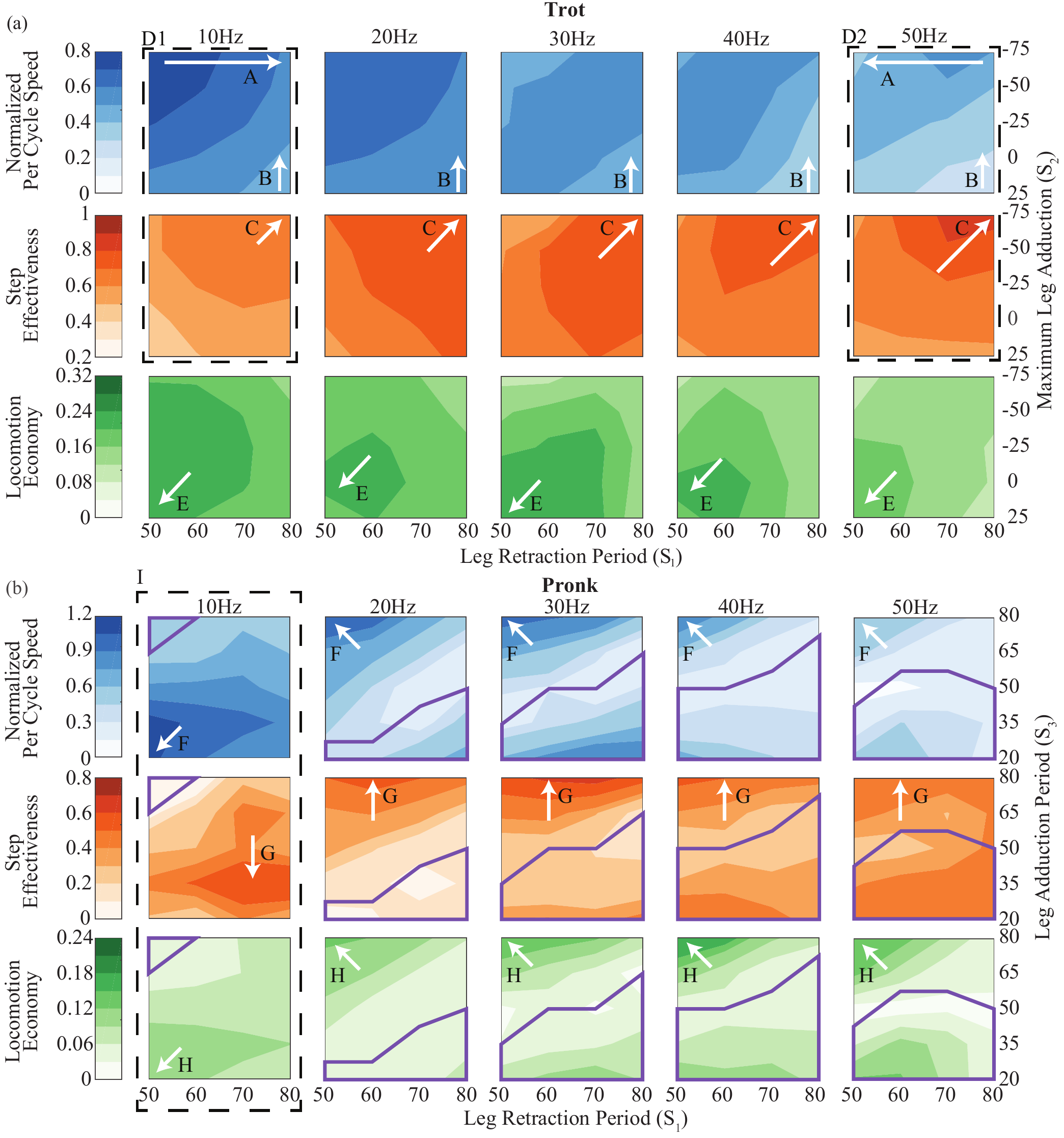}
		\vspace{-0.25cm}
		\caption{Contour plots depict the effect of the trajectory parameters, S1 ($x$-axis) and S2 ($y$-axis), on locomotion performance quantified by normalized per-cycle speed (blue), stride effectiveness (orange) and locomotion economy (green) as a function of stride frequency (\SIrange{10}{50}{\hertz}). (a) For the trot gait, the trajectory parameters are Leg Retraction Period ($x$-axis) and Maximum Leg Adduction ($y$-axis). (b) For the pronk gait, the same are Leg Retraction Period ($x$-axis) and Leg Adduction Period ($y$-axis). The purple polygons indicate regions where locomotion was backward. Labels A-H refer to points of specific interest and are discussed in the text in Sec.~\ref{sec:gait_results}.} 
		\label{fig:performance_landscape}
		\vspace{-0.75cm}
	\end{center}
\end{figure*}

The average value for each locomotion performance metric described in Sec.~\ref{sec:perf_metrics} are plotted as a function of the shape control parameters (Tab.~\ref{tab:trajdesign}) at all five tested stride frequencies (\SIrange{10}{50}{\hertz}) in Fig.~\ref{fig:performance_landscape}. We first summarize the robot's locomotion performance for the trot gait, validate hypotheses $H_1$ and $H_3$, and invalidate hypothesis $H_2$. We then summarize performance for the pronk gait and validate hypotheses $H_1$ and $H_4$. 

\subsection{Trot Gait Performance Summary}

As shown in Fig.~\ref{fig:performance_landscape}a, we are able to achieve locomotion over a wide range of speeds (\SIrange{43}{278}{\milli\meter\second^{-1}} or \SIrange{0.95}{6.17}{BL\second^{-1}}, $n=200$ trials, $N=2$ robots) by varying stride frequency and the shape control parameters. We also measure step effectiveness for the above gaits ranging from \SI{0.25}{} to \SI{0.91}{} (Fig.~\ref{fig:performance_landscape}a). In addition, we find that locomotion economy (Fig.~\ref{fig:performance_landscape}a) varies nearly four-fold (\SIrange{0.08}{0.30}{}) and shows a strong dependence on shape control parameters both within and across frequencies. The resulting cost of transport (COT) values range from \SIrange{3.33}{13.14}{}, and are some of the lowest measured on this platform \cite{goldberg2017highb,goldberg2017gait}. Finally, we note that cost of transport increases with frequency while maintaining a trot, supporting the hypothesis that the preferred gait varies as a function of running speed \cite{hildebrand1965symmetrical}. The best and worst performing trials are visualized in supplementary video S2.


\subsection{$H_1$ - Trot Gait}

For all stride frequencies, a higher leg retraction period results in increased step effectiveness (C, Fig.~\ref{fig:performance_landscape}a). Leg retraction period, however, is only positively correlated with per-cycle velocity at high stride frequencies (A, Fig.~\ref{fig:performance_landscape}a). Finally, a higher leg retraction period results in lower locomotion economy at all stride frequencies (E, Fig.~\ref{fig:performance_landscape}a). These trends support our initial hypothesis ($H_1$) that increasing leg retraction period increases step effectiveness by decreasing slipping. However, step effectiveness is only a good predictor of speed at high stride frequencies (D2, Fig.~\ref{fig:performance_landscape}a), and the two are uncorrelated at low stride frequencies (D1, Fig.~\ref{fig:performance_landscape}a). This is because the body dynamics (Fig.~S1) have a dominating effect on speed at lower stride frequencies. These dynamics, however, are attenuated at higher stride frequencies, and, therefore, speed in those regimes is largely determined by the magnitude of foot slipping \cite{goldberg2017gait}. This negative correlation between locomotion economy and leg retraction period also indicates that the energetic cost of tracking the high-velocity leg protraction might offset the benefit of mitigating leg slip. Finally, our results corroborate previous findings \cite{karssen2011optimal, haberland2011effect, karssen2015effects} that imply the existence of preferred values of leg retraction period that minimize foot slippage and economy respectively. Moreover, we find that these values are a function of the stride-frequency dependent dynamics of the robot.

\subsection{$H_2$ and $H_3$ - Trot Gait} 

For all stride frequencies, higher maximum leg adduction results in both higher step effectiveness (C, Fig.~\ref{fig:performance_landscape}a) and higher per-cycle velocity (B, Fig.~\ref{fig:performance_landscape}a). These trends refute our initial hypothesis ($H_2$) that increasing the maximum leg adduction reduces locomotion performance in terms of speed. It is likely that higher maximum leg adduction results in increased normal and frictional support, both reducing slipping and improving forward speed. 

Furthermore, increasing maximum leg adduction increases the effective leg stiffness (see Fig.~S2 and Note S3) and this likely allows for greater energy storage and return, facilitating faster locomotion and supporting our initial hypothesis ($H_3$). We suspect this is because increasing maximum leg adduction increases the relative leg stiffness for HAMR by a factor of $\sim$2 from 4.3 with zero maximum leg adduction \cite{goldberg2017gait}. The robot's relative stiffness now approaches what is observed in in animals ($\sim$10 \cite{blickhan1993similarity}) resulting in effective SLIP-like locomotion \cite{cavagna1977mechanical}. However, higher maximum leg adduction results in lower locomotion economy across all stride frequencies (E, Fig.~\ref{fig:performance_landscape}a). This suggests that increasing maximum leg adduction increases power consumption; however, this increase does not enable proportional gains in output mechanical power (i.e., forward speed) and results in less effective locomotion.  

\subsection{Pronk Gait Performance Summary} 

We find that modulating the timing between vertical and fore-aft leg motions enables locomotion over a wide range of speeds (\SIrange{-176}{236}{\milli\meter\second^{-1}} or \SIrange{-3.91}{5.24}{BL\second^{-1}}, $n=200$ trials, $N=2$ robots; blue contours in Fig.~\ref{fig:performance_landscape}b) in both forward and reverse directions. The fastest trials ($\nu>1$) are highly dynamic with long aerial and short stance phases. We also observe that step effectiveness varies from \SI{0.01}{} to \SI{0.76}{} (orange contours, Fig.~\ref{fig:performance_landscape}b). In addition, we find that locomotion economy (green contours, Fig.~\ref{fig:performance_landscape}b) varies nearly fifteen-fold (\SIrange{0.02}{0.24}{}). The resulting COT values (\SIrange{4.21}{64.84}{}) span the range from being among the lowest measured for this platform to some of the highest at each frequency. Finally, we note that actuator per-cycle energy consumption is independent of the stride frequency and the gait shape control parameters, and, as a consequence, the contour maps of $\epsilon$  mirror that of $\nu$. The best and worst performing trials are visualized in supplementary video S3.

\begin{figure*}[ht]
	\begin{center}
		\includegraphics[width=\textwidth]{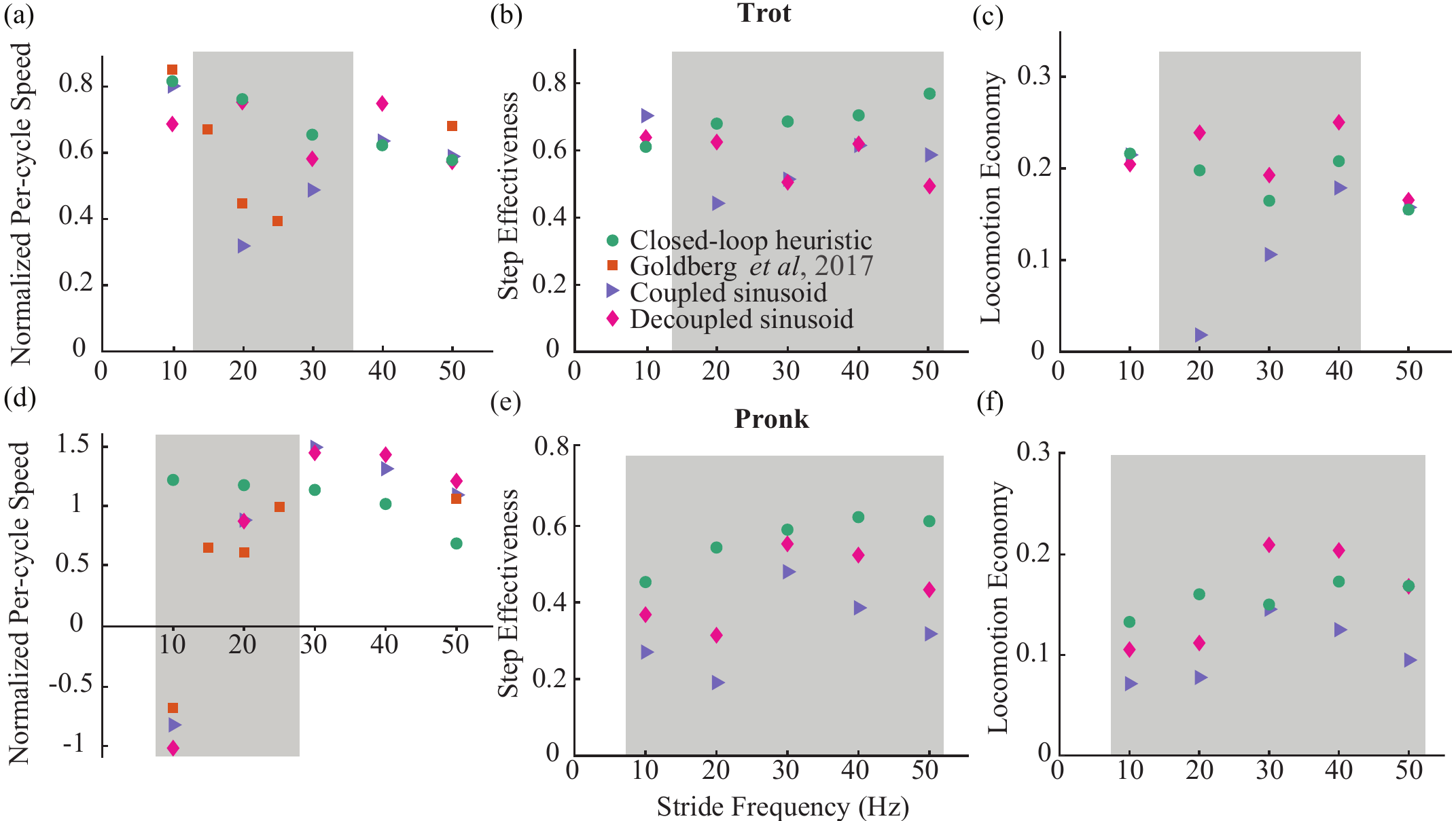}
		\caption{Plot of performance metrics -- (a,d) maximum normalized per-cycle speed, (b,e) step effectiveness, and (c,f) locomotion economy -- as a function of stride frequency (\SIrange{10}{50}{\hertz}) for the trot (top row) and pronk (bottom row) gaits. We compared performance across four different types of trajectories: closed-loop heuristic (green circles,~\ref{sec:closedlooplegexploration}), best performing trajectories from Goldberg et al. (orange squares,~\cite{goldberg2017gait}), coupled sinusoidal (blue triangles,~\ref{sec:coupledmatching}), decoupled sinusoidal (magenta diamonds,~\ref{sec:uncoupledmatching}). The gray shaded regions indicate where the closed-loop heuristic trajectories outperformed the coupled sinusoidal trajectories.} 
		\label{fig:performance_summary}
		\vspace{-0.75cm}


	\end{center}
\end{figure*}

\subsection{$H_1$ - Pronk Gait} 

We find that the lowest leg retraction period results in the highest per-cycle velocity (F, Fig.~\ref{fig:performance_landscape}b) and locomotion economy (H, Fig.~\ref{fig:performance_landscape}b) across all stride frequencies. This matches our intuition that rapid leg swing retraction during stance is key to maximizing the net forward impulse imparted to the robot. Furthermore, we do not see a clear trend in the dependence of step effectiveness on leg retraction period (G, Fig.~\ref{fig:performance_landscape}b); however, we again see that step effectiveness is a good predictor of normalized per-cycle speed at higher stride frequencies. These trends refute our initial hypothesis $H_1$ that increasing leg retraction period reduces leg slip and therefore results in improved performance. 

\subsection{$H_4$ - Pronk Gait} 

A high leg adduction period and low swing retraction period results in fast forward locomotion (F, Fig.~\ref{fig:performance_landscape}b), high step effectiveness (G, Fig.~\ref{fig:performance_landscape}b), and high locomotion economy (H, Fig.~\ref{fig:performance_landscape}b) for stride frequencies from \SIrange{20}{50}{\hertz}. Similarly, a low leg adduction period and high leg retraction period results in fast backwards (enclosed by a purple polygon) locomotion  and high locomotion economy.  Finally, intermediate values of leg adduction (independent of leg retraction) result in ineffective locomotion.    
This supports our initial hypothesis ($H_4$) that the timing between vertical and fore-aft leg motions is crucial in determining locomotion performance and direction,  and matches similar observations from previous studies \cite{hamner2013muscle, hasaneini2013optimal}. 
In contrast, we observe a reversal in the trends described above (I, Fig.~\ref{fig:performance_landscape}b) at a stride frequency of \SI{10}{\hertz} where the robots mechanical $z$-resonance results in long flight phases that favor a shorter leg adduction period.  


\section{Effective Locomotion Performance Across Dynamic Regimes}
\label{sec:gait_summary}

We analyze the best performing trials (Fig.~\ref{fig:performance_summary}) to test our final hypothesis ($H_0$) that closed-loop trajectory modulation enables high-performance locomotion across stride frequencies. Using speed as the primary metric to facilitate a comparison with previous results from \cite{goldberg2017gait}, we define the best performing trial as the one with the highest normalized per cycle speed ($\nu$) at each frequency for the trot and pronk, respectively. However, we also plot step effectiveness ($\epsilon$) and locomotion economy ($\sigma$) for the best performing trials to consider multi-dimensional robot performance. 

For the trot gait, we find that closed-loop heuristic trajectories allow the robot to maintain high speed locomotion across all stride frequencies (Fig.~\ref{fig:performance_summary}a). This is in contrast with the open-loop results from \cite{goldberg2017gait} and the coupled sinusoidal trajectories (Sec.~\ref{sec:coupledmatching}) where the robot suffers from poor performance in intermediate frequency regimes (\SIrange{15}{35}{\hertz}, supplementary video S1). However, we find that there is minimal difference in robot speed when using either the closed-loop heuristic leg trajectories or the decoupled sinusoidal trajectories (Sec.~\ref{sec:uncoupledmatching}). 
A similar trend is observed with locomotion economy (Fig.~\ref{fig:performance_summary}c); however, the closed-loop heuristic trajectories enable higher step effectiveness at all stride frequencies greater than \SI{10}{\hertz} (Fig.~\ref{fig:performance_summary}b). These results suggest that, while the shape of leg trajectories is important for effective locomotion using the trot gait in the body dynamics regime (\SIrange{15}{35}{\hertz}), the distribution of energy between the leg vertical and fore-aft motion achieved via leg shape modulation is the significant consideration at operating conditions where the dynamics are neither mechanically tuned (\SI{10}{\hertz}) nor attenuated (\SIrange{40}{50}{\hertz}). 

Similarly, we also find that closed-loop heuristic trajectories allow the robot to maintain speed across all stride frequencies (Fig.~\ref{fig:performance_summary}d) when using a pronk gait. This is in contrast with the open-loop results from \cite{goldberg2017gait}, the coupled sinusoidal trajectories, and the decoupled sinusoidal trajectories where the robot suffers from poor performance between \SIrange{5}{25}{\hertz} (supplementary video S1). On the other hand, closed-loop heuristic trajectories enable higher step effectiveness (Fig.~\ref{fig:performance_summary}e) and locomotion economy (Fig.~\ref{fig:performance_summary}f) across all stride frequencies compared to coupled input matched open-loop trajectories. This validates hypothesis $H_0$, indicating that leg trajectory modulation enables high performance locomotion across stride frequencies. 


\section{Conclusion and Future Work}
\label{sec:conclusion}

We have presented a computationally efficient framework for proprioceptive sensing and control of leg trajectories on a quadrupedal microrobot. We used this capability to explore two parametric leg trajectories designed to test a series of hypotheses investigating the influence of leg slipping, stiffness, timing, and energy on locomotion performance. This parameter sweep resulted in an experimental performance map that allowed us to select control parameters and determine a leg trajectory that maximized performance at a desired gait and stride frequency. Using these parameters, we recovered effective performance over a wide range of stride frequencies, achieving locomotion that is robust to perturbations from the robot's body dynamics \cite{jen2005robust}. 

Specifically, for the trot gait, we demonstrated that maximizing robot speed depends on minimizing slipping at high stride frequencies and leveraging favorable dynamics at low and intermediate stride frequencies. We found that the mechanism for doing either was modulating leg trajectory shape, and consequently, input energy. In addition, we were able to increase energy storage and return by modulating leg stiffness, which resulted in faster locomotion. Furthermore, we found that leg timing determined performance for the pronk gait and allowed for rapid locomotion in the forward or backwards directions.


As potential next steps towards improving the robot's state estimation, we plan to explicitly address the hybrid nature of the robot's underlying dynamics. Such an effort would require an appropriate contact sensor and a modification of the current estimation and control framework, and in principle could result in improved tracking performance. Moreover, we aim to use this low-level controller in conjunction with trajectory optimization scheme described by Doshi et al. \cite{doshi2018contact} to design feasible leg trajectories that optimize a given cost (e.g., speed, COT, etc.) at a particular operating condition. This can automate the challenging task of designing appropriate leg trajectories for a complex legged system and result in better locomotion performance. Finally, we can use this controller to ensure accurate tracking of the leg trajectories during a variety of locomotion modalities including swimming \cite{chen_controllable_2018} or climbing \cite{deRivazeaau3038} with HAMR. 

In addition the planning and control efforts discussed above, the small footprint and mass of the sensors combined with the computational efficiency of the estimation and control scheme makes our approach suitable for future implementation on the autonomous version of HAMR \cite{goldberg2018power}. We can also use the results from this work to inform future mechanical design decisions. For example, increasing the transmission resonant frequencies \cite{doshi2017phase} can increase control authority and enable improved leg trajectory control at stride frequencies higher than those   tested in this work ($>$ \SI{50}{\hertz}). Ultimately, our results suggest that HAMR could be a strong candidate platform for systematically testing hypotheses about biological locomotion such as the effect of varying leg trajectories on locomotion \cite{cruse2009principles}.

\section*{Acknowledgements}

Thank you to all members of the Harvard Microrobotics and Agile Robotics Laboratories for invaluable discussions. This work is partially funded by the Wyss Institute for Biologically Inspired Engineering. In addition, the prototypes were enabled by equipment supported by the ARO DURIP program (award $\#$W911NF-13-1-0311).

\section*{References}

\bibliographystyle{IEEEtran}
\bibliography{references}

\end{document}